# Complex-valued image denosing based on group-wise complex-domain sparsity

Vladimir Katkovnik, Mykola Ponomarenko and Karen Egiazarian

*Laboratory of Signal Processing, Tampere University of Technology, Tampere, Finland*
*E-mail: vladimir.katkovnik@tut.fi, mykola.ponomarenko@tut.fi, karen.egiazarian@tut.fi*

**Abstract**

Phase imaging and wavefront reconstruction from noisy observations of complex exponent is a topic of this paper. It is a highly non-linear problem because the exponent is a $2\pi$-periodic function of phase. The reconstruction of phase and amplitude is difficult. Even with an additive Gaussian noise in observations distributions of noisy components in phase and amplitude are signal dependent and non-Gaussian. Additional difficulties follow from a prior unknown correlation of phase and amplitude in real life scenarios. In this paper, we propose a new class of non-iterative and iterative complex domain filters based on group-wise sparsity in complex domain. This sparsity is based on the techniques implemented in Block-Matching 3D filtering (BM3D) and 3D/4D High-Order Singular Decomposition (HOSVD) exploited for spectrum design, analysis and filtering. The introduced algorithms are a generalization of the ideas used in the CD-BM3D algorithms presented in our previous publications. The algorithms are implemented as a MATLAB Toolbox. The efficiency of the algorithms is demonstrated by simulation tests.

## 1. Introduction

On many occasions processed signals have a form of complex-valued exponents $u_o = a\exp(j\varphi)$ where both phase $\varphi$ and amplitude $a$ are spatially varying.

This type of signals are typical for instance in holography, microcopy and optical metrology [1], [2], [3], [4]. A phase imaging is of a special interest in medicine and biology when a structure of specimens, for instance biological cells, is nearly invisible in usual intensity images. However, variations in thickness, density and refractive index result in variations of the phase delay of coherent monochromatic light beams. *Quantitative visualization* of these invisible phase variations by transforming them in light intensity is one of the challenging problems in optics which is fundamentally based on computational data processing [5], [6]. *Phase retrieval* is one of the computational formulations for quantitative phase imaging from intensity observations (e.g. [7]).

For complex-valued objects we meet two parallel problems: phase and amplitude imaging as these two variable define the object. Even more, it is appeared that the problem is even more complex as the phase can be treated in two very different form as the principal or interferometric phase $\varphi \in [-\pi, \pi)$ and the corresponding absolute phase as $\varphi_{abs}$. We introduce the phase-wrap operator $\mathcal{W}: \mathbb{R} \mapsto [-\pi, \pi)$, linking the absolute and principal phase as $\varphi = \mathcal{W}(\varphi_{abs})$. We also define the unwrapped phase as $\varphi_{abs} = \mathcal{W}^{-1}(\varphi)$. Note that $\mathcal{W}^{-1}$ is not an inverse operator for $\mathcal{W}$ because the latter is highly non-linear and for signals of dimension two and higher there is no one-to-one relation between $\varphi_{abs}$ and $\varphi$. The absolute phase can take arbitrary values goes beyond the interval $[-\pi, \pi)$ and it is used for measurement large magnitude variations in object parameters.

Complex-valued denoising is a reconstruction of $u_o(x)$, $a(x)$, $\varphi(x)$ and maybe $\varphi_{abs}(x)$ from the observed noisy data $z : X \to C$, where $X \subset Z^2$ is 2D grid of size $\sqrt{n} \times \sqrt{n}$, are modeled as

$$z(x) = u_o(x) + \varepsilon(x), \tag{1}$$
$$u_o(x) = a(x)e^{j\varphi(x)},$$

where $x \in X$, $u_o(x) \subset \mathbb{C}^{\sqrt{n}\times\sqrt{n}}$ is a clear complex-valued image, and $\varepsilon(x) = \varepsilon_I(x) + j\varepsilon_Q(x) \subset \mathbb{C}^{\sqrt{n}\times\sqrt{n}}$, is complex-valued zero-mean Gaussian circular white noise of variance $\sigma^2$ (i.e., $\varepsilon_I$ and $\varepsilon_Q$ are zero-mean independent Gaussian random variables with variance $\sigma^2/2$).

Phase and amplitude could be different functions of the argument $x$, for instance, with an invariant amplitude and a varying phase ("phase object") or vice versa with a varying amplitude and an invariant phase ("amplitude object"). It is quite usual that while phase and amplitude look differently similar features can be recognized in phase and amplitude images. Phase and amplitude may be highly correlated variables and a priory this correlation is unknown. It follows that naive approaches with, say, separate denoising real-valued pairs phase/amplitude or real/imaginary parts of complex-variable have no a chance to compete with intelligent algorithms developed for joint processing these variables taking into consideration the correlation between them.

Complex-valued basis functions are natural tools for processing complex-valued objects. We may mention the following well established instruments such as Fourier, windowed Fourier and Gabor transforms [8], [9], various kind of complex-valued wavelet transforms [8], [10], [11] and fresnelets [12]. The topic of sparse and redundant representations in computational imaging has attracted tremendous interest in the last years. This interest is defined by the fundamental role that low dimensional models play in many signal and image processing areas such as compression, restoration, classifications, and design of priors and regularizers, just to name a few. It is assumed in sparse imaging that there exists a basis consisting of a small number of items where image can be represented exactly or approximately with a good accuracy. This ideal basis is a priory unknown and selected from a given set of potential bases (dictionary or dictionaries) or designed from given noisy observations. This is the so-called transform domain sparsity. The compressive imaging (CI) mainly uses the same sparse approximation techniques but with the main intention to achieve the best quality/accuracy from given observations with special intention to compress data.

Recently in optics, sparsity and compressed sensing (CS) in complex domain have become a subject of study and multiple applications. Complex-valued data and operators are distinctive features of this development. Basic facts of the corresponding theory, algorithms, simulations as well as experi-mental demonstrations can be found in [13], [7]. In the works concerning the complex-valued data the phase is the most delicate and difficult issue and, in particular, the corresponding dictionary design for sparse modeling is crucial. One of the first demonstrations that the sparsity prior imposed on phase, due to the joint quadratic and total variation (TV) penalization, results in the significant improvement of complex-valued wavefront reconstruction was done in [14]. A serious accuracy improvement for wavefield reconstruction was demonstrated in [15], [16] and [17] due to a sparse BM3D modeling separate for amplitude and absolute phase in the iterative Sparse Phase Amplitude Retrieval (SPAR) algorithm. A first work on the learning dictionary based sparsity for complex-domain image filtering is presented in [18].

This paper is a further development and generalization of the techniques proposed in our recent publications [19], [20] and [21]. The ideas of these techniques are fundamentally based on group-wise sparsity and High Order Singular Value Decomposition (HOSVD). The group-wise sparsity is used in the following form. The complex-valued image is partitioned into small overlapping rectangular patches. For each patch, a group of similar patches is collected from a pre-defined neighborhood and stacked together forming 3D/4D arrays (groups). The patches in these groups allow a sparse representation just due to similarity of the patches in the groups. HOSVD is used for another purposes. It works as an instruments for design of the analysis and syntheses transforms. These transforms being complex-valued are enable to decorrelate data in the groups and in this way allow to select the main components of the approximations which are able to serve as the sparse representations. It should be emphasized that these transforms decorrelate phase and amplitude variables and in this way automatically take into consideration the correlations between these variables in the object to be reconstructed.

The novelty of this paper in comparison with the cited predecessors can be formulated as follows: 1) The non-iterative algorithms from [21] are further developed as a family of HOSVD based algorithms; 2) A new class of the iterative algorithms is developed which demonstrates a convincing improvement in the accuracy; 3) A general interpretation of the complex domain sparsity concept implemented in the algorithms is presented; 4) A thorough comparative study of the algorithms is produced. The presented non-iterative and iterative algorithms are implemented in MATLAB and presented as a toolbox: Complex Domain Image Denoising (CDID Toolbox).

The paper is concentrated on the denoising problem with an additive Gaussian noise as it is shown in Eqs.(1). However, the value of these algorithms is much wider as they can be used for smoothing and filtering in various scenarios. In particular, they can be incorporated in algorithms obtained from variational formulations. It has been demonstrated recently that denoising algorithms, designed for additive noise observations, can serve as efficient regularizers in various CI problems as "plug-and-play priors" [22], [23], [24], [25].

In what follows the paper is organized as follows. The group-wise complex domain sparsity is a topic of Section 2. The review of the BM3D based complex-valued filters is given in Section 3 where the further development is presented with introduction of the novel iterative algorithms. The flexibility of the proposed class of the algorithm allows to fit them to various applications. The extended experimental Section 4 is targeted on comparison of the algorithms based on the accuracy criteria. These experiments show that the iterative algorithm using the sparsity in the space of real/ imaginary parts of complex signals can be treated as the best in the class of the designed algorithms as well as in comparison with the state-of-the-art in the field.

## 2. Group-wise complex domain sparsity and thresholding

*2.1. Sparsity*

Following the conventional procedure in patch-based image restoration, the noisy $\sqrt{n} \times \sqrt{n}$ image $u \equiv \{u(x),\ x \in X\}$, $z \subset \mathbb{C}^{\sqrt{n} \times \sqrt{n}}$, is partitioned into small overlapping rectangular/square patches $N_1 \times N_2$ defined for each pixel of the image.

This talk is of a general nature, and $u(x)$ can be noisy image $z(x)$, true image or estimate of the true image.

Let $\mathbf{P}_x \equiv \{u(y), y \in \mathcal{P}_x \subset X\}$ denote an image patch of size $N_1 \times N_2$ defined on the domain $\mathcal{P}_x$, where the index $x \in X$ corresponds to the upper-left pixel of the patch. For each $r$-th patch (reference patch) we select $J_r$ similar patches which are closest to the reference patch $\mathbf{P}_r$. We define a 3D group $G_r \subset X$ as

$$G_r \equiv \{x \in X : \underline{b}_r \leq d(\mathbf{P}_x - \mathbf{P}_r) \leq \bar{b}_r\}, \qquad (2)$$

where $d(\mathbf{P}_x - \mathbf{P}_r)$ denotes the Euclidean distance between complex-valued patches $\mathbf{P}_x$ and $\mathbf{P}_r$, and $\underline{b}_r$, $\bar{b}_r$ are parameters controlling the desirable level of similarity in the group.

Let the matched patches $\mathbf{P}_x$, for $x \in G_r$ are stacked together to form a 3D array of the size $N_1 \times N_2 \times J_r$, denoted by $u^r$, where $J_r$ denotes a length of the array (the number of elements in $G_r$).

The transform sparsity for $u^r$ can be formalized using matrix operations:

$$\mathbf{u}^r = \mathbf{\Psi}_{\mathbf{u}^r} \theta_{\mathbf{u}^r}, \; \theta_{\mathbf{u}^r} = \mathbf{\Phi}_{\mathbf{u}^r} \mathbf{u}^r, \qquad (3)$$

where the capital $\mathbf{u}^r \in \mathbb{C}^{p_r}$, $p_r = N_1 N_2 J_r$, is a vectorized representation of the 3D tensor $u^r$ and $\theta_{\mathbf{u}^r} \in \mathbb{C}^{P_r}$ is complex-valued spectrum of the $r-th$ group.

Herein, the syntheses $\mathbf{\Psi}_{\mathbf{u}^r}$ and analysis $\mathbf{\Phi}_{\mathbf{u}^r}$ matrices (transforms, dictionaries) for $u^r$ are also complex-valued. Following the sparsity rationale it is assumed that the spectrum $\theta_{\mathbf{u}^r}$ is sparse; i.e., most elements thereof are zero. In order to quantify the level of sparsity of $\theta_{\mathbf{u}^r}$, i.e., its number of non-zero (active) elements, we use the pseudo $l_0$-norm $\|\cdot\|_0$ defined as a number of non-zero elements of the vector-argument. Therefore, following the sparsity concept approach, we design algorithms promoting low values of $\|\theta_{\mathbf{u}^r}\|_0$.

The transform domain is redundant if $P_r \geq p_r$, i.e. the analysis spectrum is of a larger dimension than the dimension of the group-image. It is a common practice to use highly redundant, overcomplete dictionaries with $P_r \gg p_r$, providing for the analysis a huge size analysis spectrum $\theta_{\mathbf{u}^r}$.

While the above speculations are standard for the concept of the sparsity, the complex domain provides a number of additional opportunities.

The complex-valued $u^r$ can be considered as a function of two pairs of real-valued variables: amplitude/phase, $a^r$, $\varphi^r$, and real/imaginary parts of $u^r$, $\text{Re}(u^r) = a^r \cos \varphi^r$, $\text{Im}(u^r) = a^r \sin \varphi^r$.

Respectively, the complex domain sparsity can be imposed in the following three different types:

(I) Complex domain sparsity treating $u^r$ as a complex-valued variable;

(II) Joint sparsity imposed of real and imaginary parts of $u^r$;

(III) Joint sparsity of amplitude and phase of $u^r$.

The complex-domain sparsity (Type I) corresponds to the analyses/synthesis in the form (3) with complex-valued $\mathbf{u}^r$. For the types II and III the image vector is of a double size and composed from the real-valued variables corresponding respectively to the definition of the type. The formulas (3), respectively for Type II and Type III, take the form

$$\begin{pmatrix} \text{Re}(\mathbf{u}^r) \\ \text{Im}(\mathbf{u}^r) \end{pmatrix} = \mathbf{\Psi}_{\mathbf{u}^r}^{\text{Re,Im}} \theta_{\mathbf{u}^r}^{\text{Re,Im}}, \ \theta_{\mathbf{u}^r}^{\text{Re,Im}} = \mathbf{\Phi}_{\mathbf{u}^r}^{\text{Re,Im}} \begin{pmatrix} \text{Re}(\mathbf{u}^r) \\ \text{Im}(\mathbf{u}^r) \end{pmatrix}, \tag{4}$$

$$\begin{pmatrix} \mathbf{a}^r \\ \varphi^r \end{pmatrix} = \mathbf{\Psi}_{\mathbf{u}^r}^{a,\varphi} \theta_{\mathbf{u}^r}^{a,\varphi}, \ \theta_{\mathbf{u}^r}^{a,\varphi} = \mathbf{\Phi}_{\mathbf{u}^r}^{a,\varphi} \begin{pmatrix} \mathbf{a}^r \\ \varphi^r \end{pmatrix}. \tag{5}$$

It is natural to believe that the items of group composed from similar patches allows a sparse representation, i.e. there is a small number of basic functions such that this array can be well approximated. Nevertheless, a proper design or selection of the analysis and syntheses dictionaries is of importance. In particular for the types II and III, we wish to use the correlation between the real-valued pairs amplitude/phase and real/imaginary parts of complex variables. Decorrelation of the signals to get an efficient sparse representation is one of the well established ideas.

Singular Value Decomposition (SVD) and High-Order Singular Value Decomposition (HOSVD) are tools which we use for decorrelation and compact representation of multidimensional data.

Let us treat a 3D group $u^r \subset \mathbb{C}^{N_1 \times N_2 \times J_r}$ as a tensor of the dimension

$N_1 \times N_2 \times J_r$. The elements of this tensor can be expressed as $u^r_{l_1,l_2,l_3}$ with $l_1 = 1, ..., N_1$, $l_2 = 1, ..., N_2$ and $l_3 = 1, ..., J_r$. In order to treat the group $u^r$ as a whole $3D$ entity, techniques on the multilinear algebra can be used in order to take into account correlations inside and between patches. It is well known that SVD is important for matrix analysis. Similarly, there are a number of various tensor decompositions as the most important ones we mention TUCKER3 and PARAFAC [26], [27], [28].

In this paper we use the HOSVD (TUCKER3) transform allowing to represent the group-tensor in the form

$$u^r = \mathbf{S}^r \times_1 \mathbf{T}_{1,r} \times_2 \mathbf{T}_{2,r} \times_3 \mathbf{T}_{3,r}, \tag{6}$$

where $\mathbf{T}_{1,r} \subset \mathbb{C}^{N_1 \times N_1}$, $\mathbf{T}_{2,r} \subset \mathbb{C}^{N_2 \times N_2}$ and $T_{3,r} \subset \mathbb{C}^{N_{J_r} \times N_{J_r}}$ are orthonormal transform matrices, $\mathbf{S}^r \in \mathbb{C}^{N_1 \times N_2 \times J_r}$ is the so-called *core tensor*, and symbols $\times_1, \times_2, \times_3$ stand for the products of the corresponding modes (variables). The matrix transform $\mathbf{T}_{1,r}$ acts with respect to the variable $l_1$ in $\mathbf{Z}^r_{l_1,l_2,l_3}$ provided that $l_2$ and $l_3$ are fixed, similar meaning have the mode transforms $\times_2 \mathbf{T}_{2,r}$ and $\times \mathbf{T}_{3,r}$ with respect to the variables $l_2$ and $l_3$.

The formula (6) defines the signal through its spectrum, it is the synthesis transform according to (3).

The analysis transform can be represented as follows

$$\mathbf{S}^r = u^r_z \times_1 \mathbf{T}^H_{1,r} \times_2 \mathbf{T}^H_{2,r} \times_3 \mathbf{T}^H_{3,r}, \tag{7}$$

where $'H'$ stands for the Hermitian transpose.

Using the formulas (6) and (7) the analysis and synthesis transforms are calculated using $3D$ groups without vectorized representations requiring explicit calculations of the high-dimension matrices $\mathbf{\Psi}_{\mathbf{u}^r}$ and $\mathbf{\Phi}_{\mathbf{u}^r}$.

For the Type II and Type III sparse representations we use $4D$ HOSVD [21].

For the sparsity Type II, we calculate the imaginary and real parts for all items of the 3D tensor $(u^r_1, u^r_2) \subset \mathbb{R}^{N_1 \times N_2 \times J_r}$ and obtain two tensors, each of dimension $N_1 \times N_2 \times J_r$. We join these two tensors in a single tensor of the dimension $N_1 \times N_2 \times J_r \times 2$, where $N_1 \times N_2$ is the dimension of the patches, $J_r$ is the length of the group (tensor) and 2 is a number of $3D$ tensors combined in the single one of the dimension $4D$.

For the sparsity Type III, we calculate amplitude and phase of the 3D tensor $(u_1^r, u_2^r) \subset \mathbb{C}^{N_1 \times N_2 \times J_r}$ and obtain two $3D$ tensors which again can be combined in a single $4D$ tensor of the dimension $N_1 \times N_2 \times J_r \times 2$.

The 4D tensors obtain in this way require 4D HOSVD for analysis and synthesis. If we will use notation $u^r$ for the introduced $4D$ tensors the above formulas can be used for the analysis and synthesis in the form

$$u^r = \mathbf{S}^r \times_1 \mathbf{T}_{1,r} \times_2 \mathbf{T}_{2,r} \times_3 \mathbf{T}_{3,r} \times_4 \mathbf{T}_{4,r}, \qquad (8)$$
$$\mathbf{S}^r = u_z^r \times_1 \mathbf{T}_{1,r}^H \times_2 \mathbf{T}_{2,r}^H \times_3 \mathbf{T}_{3,r}^H \times_4 \mathbf{T}_{4,r},$$

where the all transform matrix as well as the core tensor are real-valued, and the core tensor has a dimension $N_1 \times N_2 \times J_r \times 2$.

In the standard SVD, the spectral matrix is diagonal and its diagonal elements are singular values of the matrix. Often, a given matrix is well approximated by a small number of singular components corresponding to the dominant singular values. These truncated SVD based approximations have been extensively used in signal and image processing both to carry out denoising and to obtain low rank approximation of the original matrices.

$3D$ HOSVD applied to the complex-valued data gives complex-valued orthonormal transform matrices $\mathbf{T}_{1,r} \times_2 \mathbf{T}_{2,r} \times_3 \mathbf{T}_{3,r}$ and a complex-valued core matrix $\mathbf{S}^r$. Contrarily to the 2D case, the core tensor is not diagonal. However, as show the experiments, in our application, a small number of tensor components with large energy dominate the group representation. Thus, assuming that the smaller elements of $\mathbf{S}^r$ are linked to noise and not to essential components of the signal, the standard element-wise thresholding filtering of $\mathbf{S}^r$ can be used in order to get sparse representations for all three types of sparse coding in complex-domain.

In a similar way, $4D$ HOSVD gives non-diagonal core tensor and the element wise thresholding can be used in order to obtain sparse modeling for the corresponding input data $u^r$.

*2.2. Thresholding*

This element-wise thresholding for filtering $\mathbf{S}^r$ is used in the form:

$$\hat{\mathbf{S}}^r = \text{thresh}(\mathbf{S}^r, \delta_r), \qquad (9)$$

where thresh($\cdot$) stands for the hard-/soft-threshold.

As per rule derived in [29] we select as the universal threshold parameter

$$\delta_r = \eta\sigma\sqrt{2\log N_1 N_2 J_r}, \qquad (10)$$

where $\eta$ parameter of the algorithm is selected from experiments.

After the thresholding the filtered group data are reconstructed using the formulas (7) and (??) as

$$\hat{u}^r = \hat{\mathbf{S}}^r \times_1 \mathbf{T}_{1,r} \times_2 \mathbf{T}_{2,r} \times \mathbf{T}_{3,r}., \qquad (11)$$

$$\hat{u}^r = \hat{\mathbf{S}}^r \times_1 \mathbf{T}_{1,r} \times_2 \mathbf{T}_{2,r} \times_3 \mathbf{T}_{3,r} \times_4 \mathbf{T}_{4,r}.$$

The thresholding operator *'thresh'* is different for each type of sparsity.

For the sparsity type I, it is defined as follows:

for the hard thresholding

$$\hat{\mathbf{S}}^r = \begin{cases} \mathbf{S}^r, \text{ if } |\mathbf{S}^r| \geq \delta_r \\ 0 \text{ if } |\mathbf{S}^r| < \delta_r \end{cases}, \qquad (12)$$

and for the soft-thresholding

$$\hat{\mathbf{S}}^r = \begin{cases} (|\mathbf{S}^r| - \delta_r)_+ (\mathbf{S}^r/|\mathbf{S}^r|), \text{ if } |\mathbf{S}^r| \geq \delta_r \\ 0 \text{ if } |\mathbf{S}^r| < \delta_r \end{cases}. \qquad (13)$$

The formulas (12) and (13) should be understood as the element-wise for the tensor $\mathbf{S}^r$. They mean that the thresholding concerns only the amplitude of the signal while the phase of the input signal is preserved. The latter property is enabled by the factor $(\mathbf{S}^r/|\mathbf{S}^r|)$ in the latter formulas.

We use this definition of the thresholding because it is agreed with the following variational formulations.

Let $x, y \in \mathbb{C}^1$ and $\hat{y} = \arg\min_y(\frac{1}{2}|x-y|^2 + \alpha|y|_p)$. Then $\hat{y} = thresh(x,\alpha)$, where *thresh* is defined by the formulas (12) and (13) for $p = 0$ (hard-thresholding)

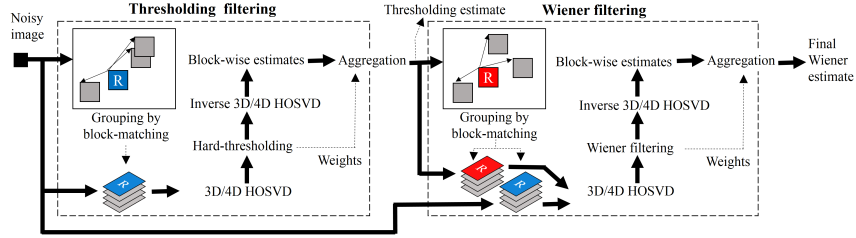

Figure 1: Flow chart of complex domain BM3D filter.

and $p = 1$ (soft-thresholding). Note that $\delta_r = \sqrt{2\alpha}$ for the hard-thresholding and $\delta_r = \alpha$ for the soft-thresholding.

For the types II and III the hard- and soft-thresholdings are applied to real valued pairs real/imaginary and amplitude/phase parts of the complex-valued variables. For these thresholds the formulas (12)-(13) take the standard forms, for the hard thresholding

$$\hat{\mathbf{S}}^r = \begin{cases} \mathbf{S}^r, \text{ if } |\mathbf{S}^r| \geq \delta_r \\ 0 \text{ if } |\mathbf{S}^r| < \delta_r \end{cases}, \qquad (14)$$

and for the soft-thresholding

$$\hat{\mathbf{S}}^r = \begin{cases} sign(\mathbf{S}^r)(|\mathbf{S}^r| - \delta_r)_+, \text{ if } |\mathbf{S}^r| \geq \delta_r \\ 0 \text{ if } |\mathbf{S}^r| < \delta_r \end{cases}. \qquad (15)$$

## 3. Algorithms

*3.1. Basic non-iterative algorithms*

Based on the introduced above concept of the complex-domain sparsity and the BM3D structure the novel algorithms for filtering of complex-valued images have been proposed with notations CD-BM3D, ImRe-BM3D and PhAm-BM3D, where CD, ImRe and PhAm refer to the corresponding sparsity types I, II, and III [21].

All these algorithms have the structure shown in Fig. 1 and composed from two successive stages: thresholding and Wiener filtering. Each of these stages include: grouping. 3D/4D HOSVD analysis, thresholding in the thresholding stage and Wiener filtering for the spectra and aggregations.

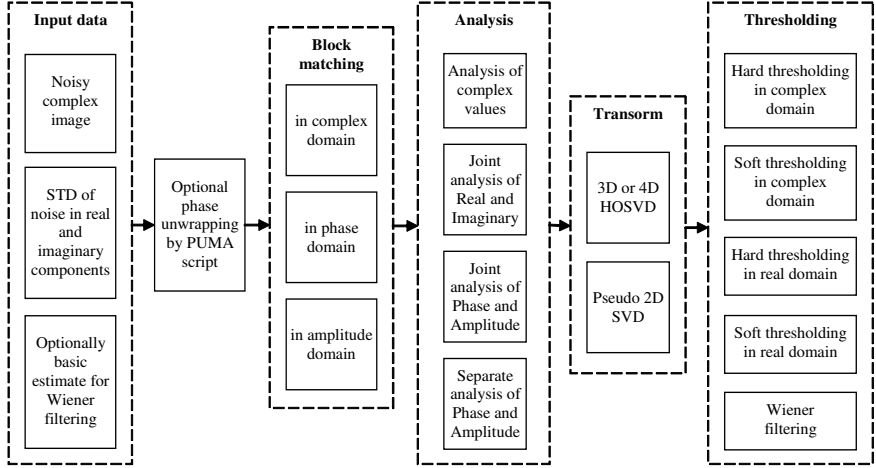

Figure 2: Structure of the introduced family of the complex domain filters

Full details of these algorithms and in particular about the specific of the Wiener filtering can be seen in [21].

*3.2. Extended set of non-iterative algorithms*

In this paper we extend the set of the above non-iterative algorithm by combination of the basic stages of these algorithms such as: 1) grouping and block-matching; 2) character of the transforms; 3) types of the spectrum analysis including the options to change the interferometric phase for the absolute one; 4) forms of thresholding (filtering spectral variables); 5) types of Wiener filtering. Joint with the corresponding parameter optimization this design results in flexible and potentially more precise algorithms.

The Fig. 2 shows the building blocks of this structural options of the available algorithms.

The columns of this figure shows the possible versions of the algorithms.

1) Grouping and block-matching can be done for complex-variable variables or for phase and amplitudes of these variables.

2) The spectrum analysis can be produced in the complex domain or in different versions of the real domains including the passage to the absolute phase by unwrapping the interferometric phase;

3) For the transform design we can use 3D-HOSVD, 4D-HOSVD or 2D SVD provided that the groups are reshaped to the matrix form.

4) The hard-/soft-thresholdings can be done in complex domain or separately for the real-valued components of the complex valued signals;

5) Wiener filtering can be included or excluded and produced in the complex or real domains.

*3.3. Iterative algorithms*

We propose a set of the novel iterative algorithms built using the discussed above non-iterative algorithms as core elements. These iterative algorithms have the structure derived in [20] from the variational formulation of the problem and shown in Table 1, where CDF (complex domain filter) is any from the non-iterative algorithms and $\alpha_t$, $\delta_t >$ are the parameters of iterations.

Table 1: ITERATIVE CDF ALGORITHMS

|   | |
|---|---|
|   | **Input**: $z \in \mathbb{C}^{\sqrt{n} \times \sqrt{n}}$ (noisy data) |
|   | Parameters: $\alpha > 0$ (regularization), |
|   | $\delta > 0$ (thresholding), |
|   | $K$ (iteration number). |
|   | Initialization: $u^0 = z$; |
|   | **Output**: $\hat{u} \in \mathbb{C}^{\sqrt{n} \times \sqrt{n}}$; |
| 1: | for $t = 1, .., K$ |
| 2: | $v^t = u^{t-1} + \alpha_{t-1}(z - u^{t-1})$; |
| 3: | $u^t = CDF(v^t, \delta_{t-1})$; |
| 4: | $end$ |
|   | $\hat{u} = u^K$. |

The thorough optimization of these iterative algorithms results in the three-step algorithms with the parameters $\alpha_{t-1}$ and $\delta_{t-1}$ varying from step-to-step.

The CDFs with thresholding are used in this optimization without Wiener filtering.

As a result of our analysis the following values are obtained for the parameters of the algorithm: $\delta_0$=0.9, $\delta_1$=0.5, $\delta_2$=0.4, $\alpha_0$=1, $\alpha_1$=0.35, $\alpha_2$=0.25. Surprisingly, these parameters enable optimization of the iterative algorithms for all types of the non-iterative CDFs. These parameters provides the best or at least acceptable performance for the most of the test-images in our experiments.

## 4. Simulation experiments

### 4.1. Test images

In this section, we present simulation results illustrating and comparing the performance of the developed algorithms. We follow the approach proposed in [21] where the complex-valued test images have correlated varying phase and amplitude. In this way we imitate a similarity of phase and amplitude images appeared in real life data. Being focussed on phase imaging we first select a phase image and use it with some scaling. The amplitude image is calculated as a function of the phase image also with some scaling. Note that the amplitudes designed in this paper are different from those in [21].

In our tests we use the six phase images shown in Fig. 3 : lena, cameraman, peppers, pattern, truncated Gauss and hills. We consider two cases: interferometric and absolute phase. The latter required the phase unwrapping while for the first one the algorithms operate with the phases restricted to the interval $[-\pi, \pi)$ and do not require unwrapping. In our tests the interferometric phases are scaled to the interval $[0, \pi/2]$. The amplitudes are scaled in a such way that the minimum and mean values are equal to 0.5 and 1.0, respectively.

The interferometric phase and the corresponding amplitude images are shown in Fig. 3.

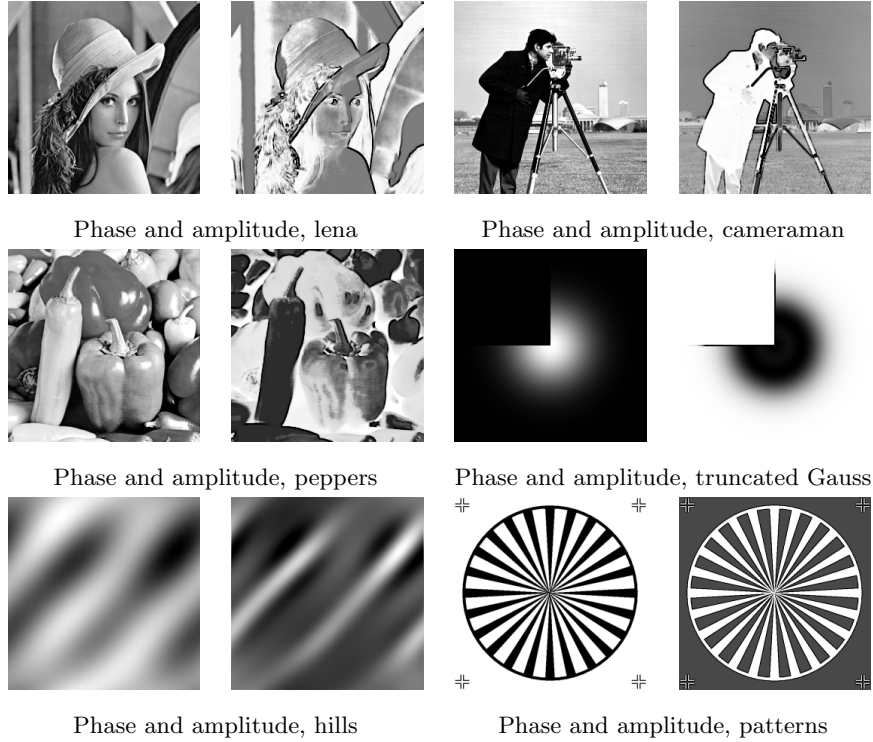

Figure 3: Complex domain test-images

The truncated Gauss and hills images are used also as the absolute phases. In this case the truncated Gauss phase is scalled to the interval $[0, 16\pi]$ radians and the hills phase is scalled to the interval $[0, 20\pi]$ radians. For the amplitude images, the minimum and mean values are 0.5 and 1.0, respectively.

4.2. Noise modeling

Following to [21], the standard deviation of the noise is calculated according to the formula

$$\sigma = \sigma_{\varphi_z} \cdot mean_x(a_o(x))\sqrt{2}, \quad (16)$$

where $mean_x(a_o(x))$ is the mean value of the amplitude and $\sigma_{\varphi_z}$ is the standard deviation of the noise in the noisy observed phase [1]. This phase scaling is introduced in order to control the noise level for the phase as the variable of the main interest in this paper and to make comparable the results of the experiments with different images.

---
[1] Note, that the factor $\sqrt{2}$ is lost in [21].

The experiments are produced for the set $\sigma_{\varphi_z} = \{0.05, 0.1, 0.2, 0.3, 0.5, 0.9\}$. The largest $\sigma_{\varphi_z}$ corresponds to the very noisy observations.

It can be proved by linearization of $z(x) = a_z(x)\exp(i\varphi_z(x))$ produced for small $\sigma$, i.e. for small random components in $a_z(x)$ and $\varphi_z(x)$, that the expectations and variances of amplitude and phase of the observed $z(x)$ are such that

$$E\{a_z(x)\} \simeq a_o(x),\ \sigma_{a_z}^2(x) = var(a_z(x)) \simeq \sigma^2/2, \qquad (17)$$
$$E\{\varphi_z(x)\} \simeq \varphi_o(x),\ \sigma_{\varphi_z}^2(x) = var(\varphi_z(x)) \simeq \sigma^2/(2a_o^2(x)).$$

Thus, experiments with varying $\sigma^2$ calculated according to (16) approximately allow to control the noise level in the phase $\sigma_{\varphi_z}^2$.

4.3. Accuracy criteria

The following performance criteria are used for evaluation of the reconstruction accuracy for the complex-valued signals.

For the interferometric phase, it is the peak signal-to-noise ratio ($PSNR$):

$$PSNR_\varphi = 10\log_{10}\frac{n(2\pi)^2}{||\mathcal{W}(\hat{\varphi}_o - \varphi_o)||_2^2}[dB], \qquad (18)$$

where $\hat{\varphi}_o$ and $\varphi_o$ are the phase reconstruction and the true phase, respectively; $n$ is the image size in pixels; the phase wrapping operator $\mathcal{W}$ is used in order to eliminate the phase shifts in errors multiple to $2\pi$ [18]. The factor $(2\pi)^2$ in the numerator of (18) stays for the squared maximum value of the interferometric phase.

The reconstruction accuracy for the amplitude $a_o$ is characterized by $PSNR$ as:

$$PSNR_{ampl} = 10\log_{10}\frac{n\max(a_o(x))^2}{||a_o - \hat{a}_o||_2^2}[dB]. \qquad (19)$$

We unwrap the estimated interferometric phase with the PUMA algorithm [30] in order to get estimates of the true absolute phase $\varphi_{o,abs}$. The accuracy of the absolute phase reconstruction is measured by the root-mean-squared error (RMSE):

$$RMSE_{\varphi_{abs}} = \sqrt{\frac{1}{n}||\varphi_{o,abs} - \hat{\varphi}_{o,abs} + \Delta_\varphi||_2^2}, \qquad (20)$$

where a scalar $\Delta_\varphi$ compensates an invariant shift in the absolute phase estimation multiple to $2\pi$ which can appear due to the unwrapping procedure. It is calculated as

$$\Delta_\varphi = 2\pi \left[ \frac{\overline{\hat{\varphi}}_{0,abs} - \overline{\varphi}_{0,abs}}{2\pi} \right]$$

Here $[\cdot]$ stands for the integer part of the argument and the hat $'-'$ means the mean value of the variable.

$RMSE$ for the amplitude reconstruction is defined as

$$RMSE_a = \sqrt{\frac{1}{n}||a_o - \hat{a}_o||_2^2}. \qquad (21)$$

Signal-to-noise ratio (SNR) for the complex-valued reconstruction of $u_o$ and the absolute phase estimate are calculated as

$$SNR_c = 10\log_{10} \frac{||u_o - \bar{u}_o||_2^2}{||u_o - \hat{u}_o||_2^2} [dB], \qquad (22)$$

$$SNR_{\varphi_{abs}} = 10\log_{10} \frac{||\varphi_{o,abs} - \bar{\varphi}_{o,abs}||_2^2}{||\varphi_{o,abs} - \hat{\varphi}_{o,abs} + \Delta_\varphi||_2^2} [dB] \qquad (23)$$

where $u_o$ stands for the true complex-valued object.

Note that $SNR_c$ evaluates the accuracy in phase and amplitude simultaneously.

### 4.4. Compared algorithms

The thorough experimental analysis showed that for the selected test-images the best performance was demonstrated by the CDID algorithms shown in the first four lines in Table 2 .

ImRe-BM3D HT, ImRe-BM3D WI and ImRe-BM3D IT mean the following versions of the ImRe-BM3D algorithms, respectively: with the hard-thresholding no Wiener filtering, with the hard-thresholding and the following Wiener filtering, and the iterative ImRe-BM3D where ImRe-BM3D is used with the hard-thresholding and no Wiener filtering. Iterative CD-BM3D also uses the CD-BM3D with the hard thresholding and without Wiener filtering.

Table 2: Compared algorithms

| # | Algorithm | Abbreviation |
|---|---|---|
| 1 | ImRe-BM3D with hard-thresholding | ImRe-BM3D HT |
| 2 | ImRe-BM3D with hard-thresholding and Wiener filtering | ImRe-BM3D WI |
| 3 | Iterative ImRe-BM3D | ImRe-BM3D IT |
| 4 | Iterative CD-BM3D | CD-BM3D IT |
| 5 | Windowed Fourier Transform | WFT |
| 6 | Sparse learning dictionary | SpInPhase |
| 7 | Iterative Sparse Phase and Amplitude Retrieval | SPAR |

For the comparative analysis we use the state-of-the-art algorithms for complex-valued images: WFT [31],[9], SpInPhase [18], and SPAR [15], [16], [17].

We wish to compare seven algorithms shown in Table 2 for the set of test-images from Fig. 3 and with various level of the additive noise in observations. In order to make this multidimensional analysis compact and transparent, instead of the mean values of PSNR or RMSE conventionally used in this sort of publications on image reconstruction, we exploit the statistical box-plots.

Let $PSNR_\varphi(k,l,m)$ be a set of $PSNR_\varphi$ calculated for $k \in K$, $l \in L$, $m \in M$, where $K$, $L$ and $M$ denote the sets of the algorithms, the noise standard deviations (six different values), and six test-images, respectively.

The best algorithm for each test image and for each noise standard deviation is defined as

$$mPSNR_\varphi(l,m) = \max_K PSNR_\varphi(k,l,m). \qquad (24)$$

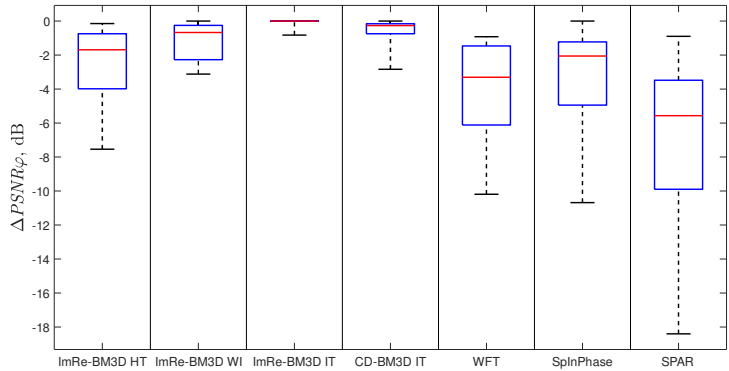

Figure 4: Box-plots of $\Delta PSNR_\varphi$ for compared algorithms

We compare the algorithms with this best result using the differences between the corresponding PSNRs:

$$\Delta PSNR_\varphi(k,l,m) = PSNR_\varphi(k,l,m) - mPSNR_\varphi(l,m). \qquad (25)$$

The box-plots being depicted for each algorithm (each $k$) gives comparative statistics with respect to two other indices $l$ and $m$, i.e. for test-images and for noise standard deviations (see Fig. 4). The upper and lower edges of the rectangle boxes correspond to 25% and 75% quantiles of $\Delta PSNR_\varphi(k,l,m)$ distributions, respectively. Dotted lines marks maximal and minimal values of $\Delta PSNR_\varphi$ for each algorithm. The horizontal line inside of the box (red in color images) is the median value of these $\Delta PSNR_\varphi$.

Let us analyze the results shown in Fig. 4. It is well seen that ImRe-BM3D IT provides the maximal values for $PSNR_\varphi$ in the overwhelming majority of the experiments. The box in this box-plot is very narrow, just a horizontal line, and the minimum values of $\Delta PSNR_\varphi$ is not much below the box. Thus, the results are very compact and close to the best possible value. ImRe-BM3D IT is outperforming ImRe-BM3D WI in average on 2 $dB$ (in some cases up to 3 dB) and outperforming ImRe-BM3D HT also in average on 2 dB (in some cases up to 7 dB). The advantage of ImRe-BM3D IT with respect to other algorithms even more valuable. Methods WFT, SpInPhase and SPAR are below than ImRe-BM3D IT of about on 2-5 dB and in some cases more than on 10 dB.

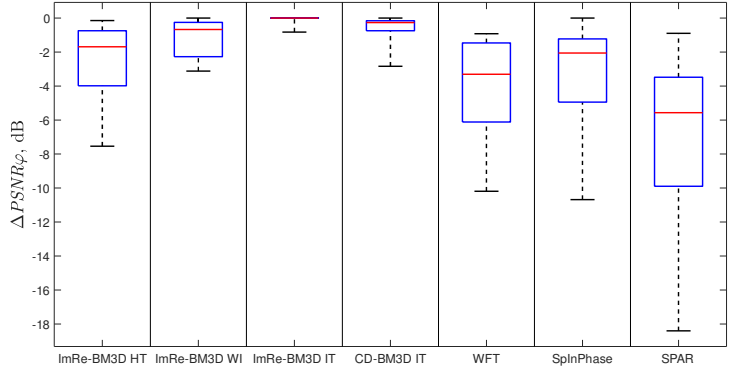

Figure 5: Box-plots of $\Delta PSNR_\varphi$ for compared algorithms

A similar analysis has been produced based on other criteria. In particular, for $SNR_c$ of the complex-valued reconstructions we can introduce the best $SNR_c$ values as

$$mSNR_c(l,m) = \max_K SNR_c(k,l,m).$$

The corresponding comparison are produced using the differences:

$$\Delta SNR_c(k,l,m) = SNR_c(k,l,m) - mSNR_c(l,m).$$

The results are shown in Fig. 5.

The comparison of $\Delta SNR_c$ again is essentially in favor of ImRe-BM3D IT.

A more specific phase accuracy comparison can be seen in Table 3 and Table 4 for $\sigma_\varphi = 0.1$.

In the most cases ImRe-BM3D IT demonstrate the best or nearly the best accuracy. For truncated Gauss and hills with large smooth areas ImRe-BM3D IT outperforms ImRe-BM3D WI more than on 2 dB. Here we wish to note than all methods from the CDID set significantly outperform WFT, SpInPhase and SPAR.

Table 3: $PSNR_\varphi$ for reconstruction of the interferometric phase, dB

| Image | Noisy | ImRe-BM3D HT | ImRe-BM3D WI | ImRe-BM3D IT | CD-BM3D IT | WFT | SpIn Phase | SPAR |
|---|---|---|---|---|---|---|---|---|
| Lena | 35.5 | 43.4 | 43.5 | **43.7** | 43.4 | 42.8 | 41.2 | 42.5 |
| Cameraman | 35.4 | 41.5 | **41.7** | 41.6 | 40.6 | 40.4 | 39.7 | 40.6 |
| Peppers | 34.6 | 43.1 | 43.3 | **43.6** | 43.4 | 42.2 | 41.0 | 39.5 |
| Gauss | 35.7 | 51.2 | 53.2 | **55.6** | 54.8 | 47.7 | 54.1 | 42.0 |
| Hills | 35.1 | 49.8 | 50.7 | 52.9 | **53.0** | 48.0 | 51.1 | 48.4 |
| Patterns | 34.5 | 43.9 | 44.4 | **45.1** | 45.0 | 40.6 | 37.8 | 40.8 |

Table 4: $SNR_c$ for interferometric phases, dB

| Image | Noisy | ImRe-BM3D HT | ImRe-BM3D WI | ImRe-BM3D IT | CD-BM3D IT | WFT | SpIn Phase | SPAR |
|---|---|---|---|---|---|---|---|---|
| Lena | 9.4 | 17.8 | 18.1 | **18.3** | 18.0 | 12.8 | 15.8 | 12.8 |
| Cameraman | 11.9 | 19.8 | **20.1** | **20.1** | 19.2 | 15.1 | 18.1 | 15.1 |
| Peppers | 10.9 | 19.5 | 19.7 | **19.9** | 19.7 | 14.7 | 17.5 | 13.9 |
| Gauss | 7.9 | 21.9 | 24.3 | **27.2** | 26.7 | 10.3 | 25.8 | 10.5 |
| Hills | 9.6 | 23.9 | 24.7 | **26.7** | 26.7 | 13.8 | 23.9 | 14.2 |
| Patterns | 14.7 | 23.9 | 24.4 | **25.1** | 25.0 | 18.5 | 18.6 | 17.7 |

A visual demonstration of phase imaging is presented in Figs 6, 7, 8 for pattern, lena and truncated Gauss test-images.

Fig. 6 shows results for the patterns test-image, $\sigma_\varphi = .3$. ImRe-BM3D IT outperforms the other methods at least on 4 dB and provides essentially better visual quality.

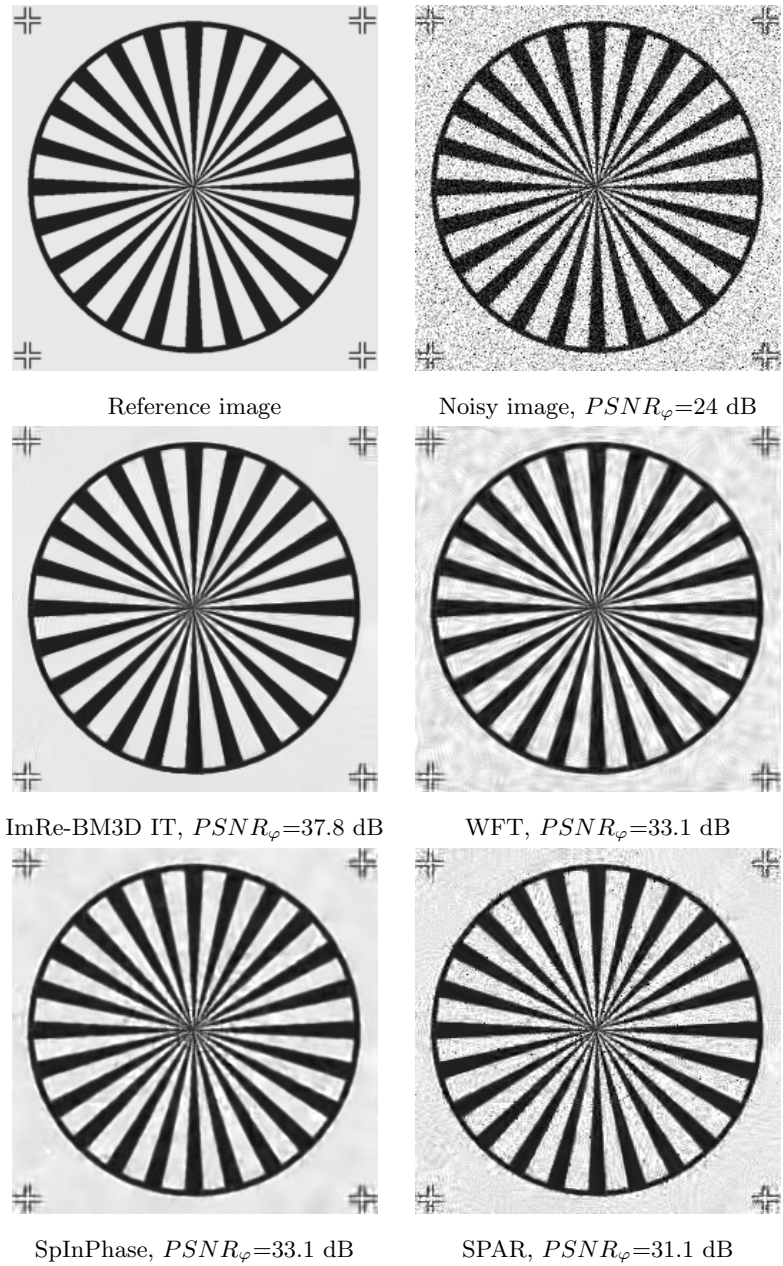

Figure 6: Denoising of test-image patterns, $\sigma_\varphi$=0.3

In Fig. 7, we can see results for the lena test-image, $\sigma_\varphi = .3$. In this case ImRe-BM3D IT outperforms SpInPhase more then on 1 dB and in the comparison with the other algorithms the improvement even higher. Definitely, the best sharpness and better imaging of edges and fine details are provided by ImRe-BM3D IT.

In Fig. 8 we show 3D imaging of the truncated Gauss phase, the observed noisy phase is extremely noisy, $\sigma_\varphi=0.2$. The comparison of ImRe-BM3D IT versus the best counterpart SpInPhase is definitely in favor of ImRe-BM3D IT both visually and numerically.

At the end of this section in Fig. 9 we show $PSNR_\varphi$ as a function of $\sigma_\varphi$. These curves are calculated as an average over six test-images. It is clearly seen that ImRe-BM3D IT provides always better accuracy with the advance about 2-3 dB.

### 4.4.1. Absolute phase test-images

For these experiments we selected truncated Gauss and hills test-phase images scalled to the intervals $[0, 16\pi]$ and $[0, 20\pi]$, respectively. These absolute phase images have a large smooth ares allowing good approximations. However, the observations are defined not by the absolute phase but the corresponding wrapped phases shown in Fig. 10, where the fringes typical for wrapped phases explain why these quite smooth absolute phase images are so difficult for denoising. Any small error in the area of the fringe means an about $2\pi$ radian error for the wrapped phase and even much larger error for the absolute phase. As a result the accuracy criteria are very sensitive to random errors in even a small number of isolated pixels. In order to decrease the sensitivity of the criterion values to the random noise in observations in our experiments we use the Monte-Carlo statistical modeling with averaging over 10 runs with independently generated noise.

Tables 5 and 6 contain values of $SNR_{\varphi_{abs}}$ for the both absolute phase test-images. In majority situations ImRe-BM3D IT provides the better denoising as compared with methods. However for truncated Gauss with small $\sigma_\varphi = 0.05$

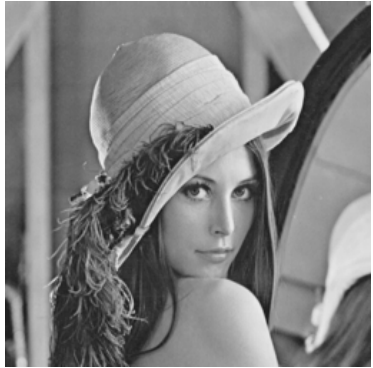
Reference image
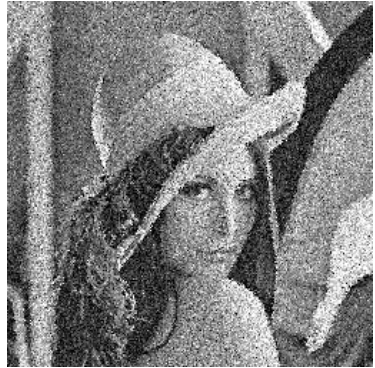
Noisy image, $PSNR_\varphi$=25.1 dB
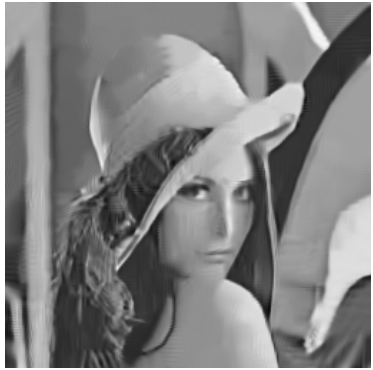
ImRe-BM3D IT, $PSNR_\varphi$=38.1 dB
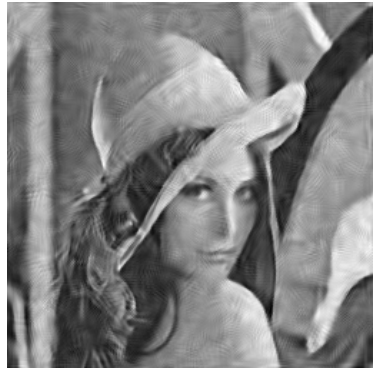
WFT, $PSNR_\varphi$=36.8 dB
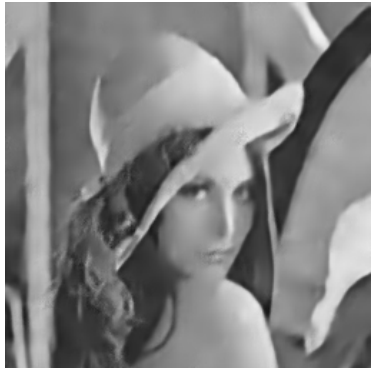
SpInPhase, $PSNR_\varphi$=37.0 dB
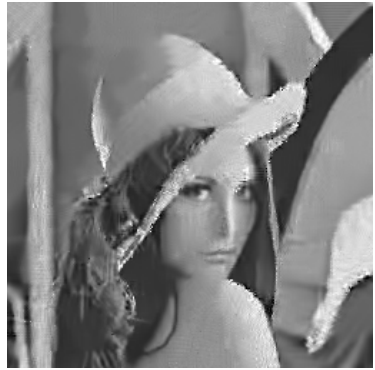
SPAR, $PSNR_\varphi$=35.6 dB

Figure 7: Denoising of test-image lena, $\sigma_\varphi$=0.3

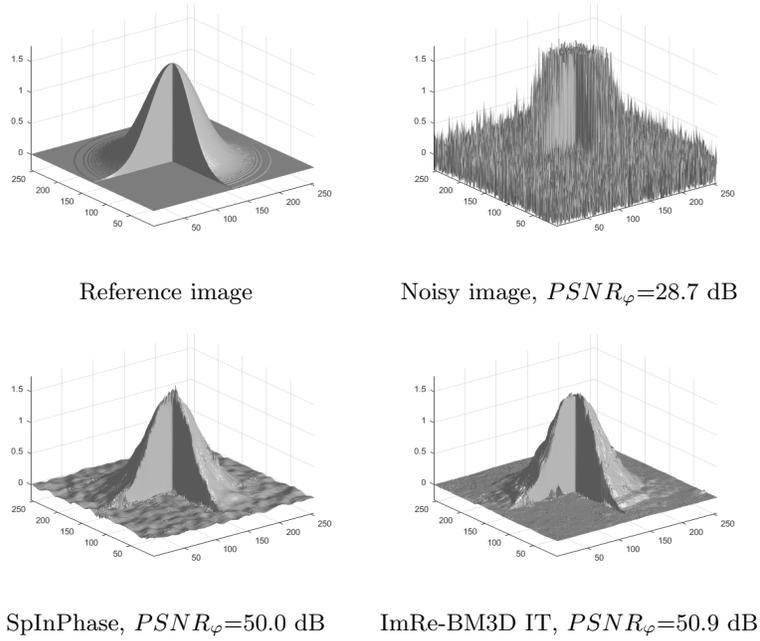

Figure 8: Results of denoising of test-image truncated Gauss, $\sigma_\varphi$=0.2

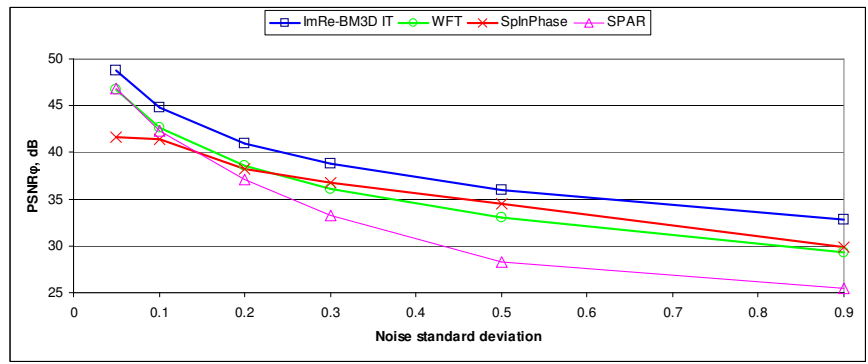

Figure 9: Averaged $PSNR_\varphi$ for interferometric phases

SpInPhase outperforms ImRe-BM3D IT on 2.5 dB.

Tables 7 and 8 contain values of $SNR_c$ for the same images. ImRe-BM3D IT surrenders to SpInPhase only for $\sigma = 0.7$ (0.2 $dB$) and for $\sigma = 0.9$ (0.5 dB), Table 10, outperforming this method as well as others up to 5 dB.

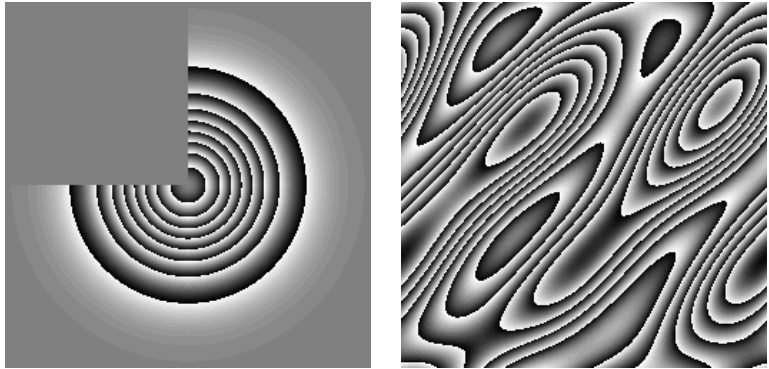

Figure 10: Wrapped phases for truncated Gauss and hills test-images

Table 5: Averaged $SNR_{\varphi_{abs}}$ (10 experiments) for truncated Gauss, dB

| $\sigma_\varphi$ | Noisy image | ImRe-BM3D IT | WFT | SpInPhase | SPAR |
|---|---|---|---|---|---|
| 0.05 | 33.1 | 43.0 | 35.2 | **45.5** | 33.2 |
| 0.1 | 30.7 | **37.4** | 35.0 | 30.9 | 31.2 |
| 0.2 | 25.8 | **33.5** | 31.0 | 30.4 | 27.7 |
| 0.3 | 14.2 | **36.0** | 28.8 | 32.9 | 17.9 |
| 0.5 | 7.2 | **29.8** | 19.9 | 20.3 | 7.3 |
| 0.9 | 3.3 | **20.2** | 16.9 | 15.5 | 3.4 |

A significant difference between the results for criteria $SNR_c$ and $SNR_{\varphi_{abs}}$ is explained by random errors in phase unwrapping by the PUMA algorithm. These errors lead to impulsive noise appeared in absolute phase reconstructions. In particular, it results in the poor performance of SPAR for high-level noise, what can be seen in Fig. 11, where the box-plot for SPAR is lower and of much larger size than for other algorithms. Remind that SPAR implements the sparsity for the absolute phase and uses the phase unwrapping on each iteration.

An example of a 3D visualization of comparative performance of SpInPhase and ImRe-BM3D can be seen in Fig. 12. The results are given for the very noisy data with $\sigma_\varphi = 0.9$. ImRe-BM3D demonstrates a quite good noise suppression for homogeneous regions outperforming SpInPhase, the closest competitor for the case, more than on 7 dB.

Table 6: Averaged $SNR_{\varphi_{abs}}$ (10 experiments) for hills, dB

| $\sigma_\varphi$ | Noisy image | ImRe-BM3D IT | WFT | SpInPhase | SPAR |
|---|---|---|---|---|---|
| 0.05 | 47.8 | 48.9 | **49.0** | 45.6 | 48.8 |
| 0.1 | 41.7 | 46.2 | 46.2 | **46.4** | 44.8 |
| 0.2 | 35.4 | **44.7** | 44.4 | 44.5 | 41.3 |
| 0.3 | 31.3 | **43.4** | 43.0 | 43.3 | 37.4 |
| 0.5 | 26.1 | **41.2** | 40.6 | 41.1 | 32.0 |
| 0.9 | 21.6 | 37.8 | 36.9 | **38.1** | 29.4 |

Table 7: Averaged $SNR_c$ (10 experiments) for truncated Gauss, dB

| $\sigma_\varphi$ | Noisy image | ImRe-BM3D IT | WFT | SpInPhase | SPAR |
|---|---|---|---|---|---|
| 0.05 | 20.6 | **30.0** | 19.0 | 23.7 | 23.9 |
| 0.1 | 14.6 | **27.0** | 18.3 | 24.9 | 18.5 |
| 0.2 | 8.6 | **23.9** | 16.8 | 22.2 | 12.7 |
| 0.3 | 5.1 | **21.9** | 15.4 | 20.5 | 9.3 |
| 0.5 | 0.6 | **19.0** | 13.2 | 17.7 | 5.2 |
| 0.9 | -4.5 | **15.1** | 9.9 | 9.8 | 1.5 |

Table 8: Averaged $SNR_c$ (10 experiments) for hills, dB

| $\sigma_\varphi$ | Noisy image | ImRe-BM3D IT | WFT | SpInPhase | SPAR |
|---|---|---|---|---|---|
| 0.05 | 23.3 | **26.2** | 20.6 | 23.0 | 24.6 |
| 0.1 | 17.3 | **23.4** | 19.6 | 23.0 | 19.5 |
| 0.2 | 11.3 | **21.3** | 18.2 | 21.0 | 14.9 |
| 0.3 | 7.7 | **19.8** | 17.0 | **19.8** | 11.7 |
| 0.5 | 3.3 | 17.3 | 15.0 | **17.5** | 7.5 |
| 0.9 | -1.8 | 13.8 | 12.0 | **14.3** | 4.1 |

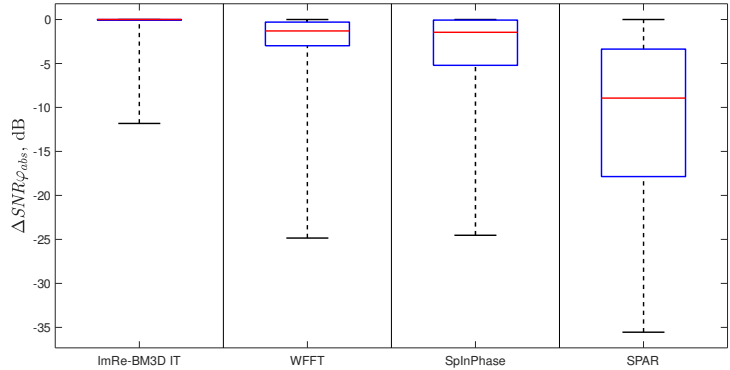

Figure 11: $\Delta SNR_{\varphi_{abs}}$ for compared methods, absolute phases

4.4.2. Comparison with the tests in [20]

The CD-BM3D algorithm with the hard-thresholding has been proposed in [19] and its iterative version (CD-BM3D IT) in [20]. Both these algorithms are included in the set of the algorithms developed in this paper. The phase test-images studied in [20] are considered with invariant amplitude and the algorithms have been tuned for these test-images. These algorithms demonstrated the uniform advantage with respect to the WFT and SpInPhase algorithms.

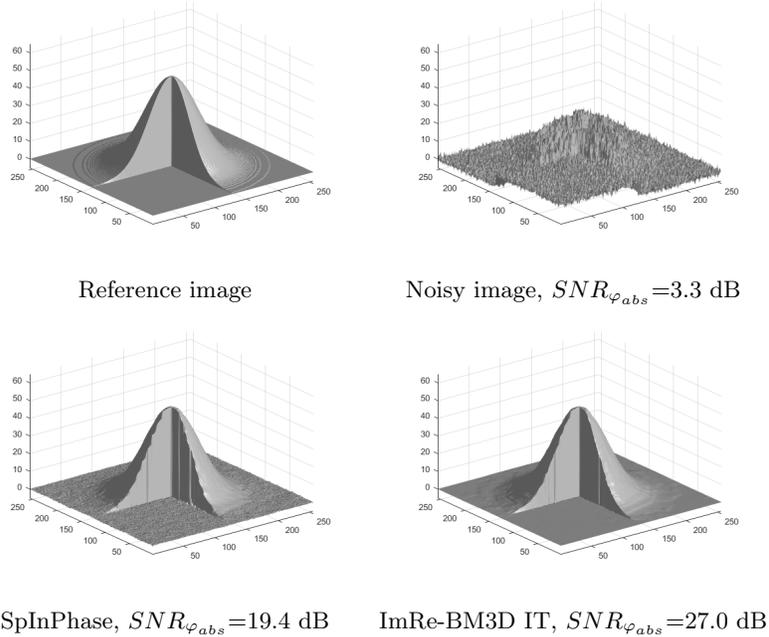

Reference image      Noisy image, $SNR_{\varphi_{abs}}$=3.3 dB

SpInPhase, $SNR_{\varphi_{abs}}$=19.4 dB      ImRe-BM3D IT, $SNR_{\varphi_{abs}}$=27.0 dB

Figure 12: Results of absolute phase reconstruction for truncated Gauss, $\sigma_\varphi$=0.9

In this section we apply ImRe-BM3D IT to the test images in [20]. The results of comparison shown in Tables 9 and 10 and can be summarized as follows. For the absolute phase reconstruction the algorithms from [20] are uniformly better. However, mainly the difference is not essential, even for some cases, for instance for Shear Plane, ImRe-BM3D IT demonstrates a serious advantage over CD-BM3D and CD-BM3D IT what can be explained by the discussed above errors in the phase unwrapping appeared due fringe discontinuous. More suprisingly is the advantage of ImRe-BM3D IT for cameraman and lena test-images.

Overall, we show these results with the only goal to demonstrate that the algorithm ImRe-BM3D IT positioned as the best in this paper can be treated as an universally good instrument which is applicable in various situations.

However, this conclusion cannot be an absolute statement as for particular applications different algorithms from our set with a special tuning may appear

Table 9: Absolute phase reconstruction for test-images from [20], $PSNR_\varphi$ dB

| Surf. | $\sigma$ | BM1 | BM2 | SpInPhase | WFT | ImRe-BM3D IT | ImRe-BM3D IT with $\delta_0$=1 |
|---|---|---|---|---|---|---|---|
| Trunc. Gauss | .1 | 50.42 | 50.94 | 48.06 | 47.56 | 50.1 | **51.2** |
| | .3 | 43.86 | **44.44** | 42.05 | 40.25 | 43.3 | 44.3 |
| | .5 | 39.66 | 40.00 | 38.76 | 36.69 | 39.6 | **40.5** |
| | .7 | 36.57 | 37.46 | 36.08 | 34.23 | 37.1 | **37.8** |
| | .9 | 33.86 | 33.89 | 33.47 | 32.24 | 35.0 | **35.2** |
| Shear plane | .1 | 56.90 | 58.17 | 58.01 | 48.25 | 57.1 | **60.1** |
| | .3 | 48.73 | 49.44 | 49.02 | 40.93 | 48.9 | **51.4** |
| | .5 | 44.37 | 46.66 | 43.65 | 37.36 | 45.8 | **47.7** |
| | .7 | 41.46 | 43.97 | 40.15 | 35.06 | 43.6 | **44.9** |
| | .9 | 39.05 | 41.78 | 33.65 | 33.17 | 42.0 | **42.8** |
| Sinus cont. | .1 | 56.65 | **57.98** | 56.26 | 43.87 | 54.1 | 54.1 |
| | .3 | 47.09 | **48.54** | 47.92 | 35.84 | 45.7 | 46.0 |
| | .5 | 42.25 | **44.05** | 43.22 | 32.09 | 41.5 | 41.3 |
| | .7 | 38.52 | **40.85** | 40.68 | 29.41 | 38.8 | 38.0 |
| | .9 | 35.48 | **37.86** | 36.29 | 27.29 | 36.8 | 35.4 |
| Sinus disc. | .1 | 55.53 | **56.40** | 48.08 | 43.87 | 51.5 | 53.8 |
| | .3 | 45.85 | **46.34** | 44.16 | 35.84 | 42.6 | 44.6 |
| | .5 | 41.33 | **42.00** | 39.54 | 32.09 | 38.2 | 40.2 |
| | .7 | 38.00 | **38.72** | 36.08 | 29.41 | 35.6 | 37.4 |
| | .9 | 35.2 | **36.34** | 33.93 | 27.29 | 33.7 | 35.4 |
| Mount. | .1 | 46.03 | **46.84** | 46.20 | 46.81 | 45.5 | 46.8 |
| | .3 | 40.99 | **41.25** | 40.32 | 39.54 | 39.3 | 39.8 |
| | .5 | 36.73 | **37.04** | 36.84 | 35.97 | 35.5 | 35.7 |
| | .7 | 33.86 | **34.37** | 33.20 | 33.51 | 32.9 | 33.0 |
| | .9 | 31.54 | **32.31** | 32.25 | 31.63 | 30.9 | 31.0 |

Table 10: Interferomatric phase reconstruction in comparison with the algorithms from[20], $PSNR_\varphi$ dB

| Surf. | $\sigma$ | BM1 | BM2 | SpInPhase | WFT | ImRe-BM3D IT |
|---|---|---|---|---|---|---|
| Lena | .1 | 45.66 | 46.17 | 41.43 | 45.42 | **46.6** |
| | .3 | 39.64 | 40.29 | 38.48 | 39.16 | **40.7** |
| | .5 | 36.89 | 37.67 | 36.73 | 36.30 | **38.1** |
| | .7 | 35.02 | 35.87 | 34.44 | 34.42 | **36.4** |
| | .9 | 33.71 | 34.55 | 34.14 | 32.93 | **35.1** |
| Cameraman | .1 | 44.06 | 44.53 | 41.00 | 43.79 | **45.3** |
| | .3 | 39.02 | 39.56 | 37.89 | 38.00 | **39.9** |
| | .5 | 36.30 | 37.04 | 35.97 | 35.35 | **37.5** |
| | .7 | 34.41 | 35.25 | 34.17 | 33.59 | **35.9** |
| | .9 | 33.11 | 33.92 | 33.28 | 32.21 | **34.7** |

to be much better.

*4.5. Computational complexity*

In our experiments we use: 64-bit Windows 7, Intel(R) Core(TM) i7-4790 3.6 GHz, 16 GB RAM, MATLAB 2016a.

The computational complexity is characterized by the computational time. The comparison of the algorithms is shown in Table 11, where approximate values of the time required for processing a 256×256 image are shown.

All methods except of SPAR use only MATLAB code. SPAR uses also internal mex-files (dlls) for acceleration of block matching and transforms calculation.

WFT is the fastest algorithm. The second place belongs to CD-BM3D. The algorithms PhAm-BM3D, Separate PhAm-BM3D (separate phase and ampli-tude filtering based on BM3D equipped with HOSVD), ImRe-BM3D are equivalent in the required time for processing a single image. ImRe-BM3D IT composed from the three iterations of ImRe-BM3D HT naturally requires threefold computational time of ImRe-BM3D HT. SpInPhase and SPAR are outsiders in this competition.

Table 11: Comparative computational time of the algorithms in sec., 256x256 images

| CD-BM3D | PhAm-BM3D | Separate PhAm-BM3D | ImRe-BM3D HT | ImRe-BM3D WI | ImRe-BM3D IT | WFT | SpIn Phase | SPAR |
|---|---|---|---|---|---|---|---|---|
| 21.4 | 26.8 | 27.7 | 26.1 | 52.3 | 78.4 | 19.2 | 134.3 | 163.7 |

There are several ways to reduce the computational complexity of BM3D based algorithms by tuning their parameters:

1) Decreasing a size of the search area for similar patches (for block matching). The default size in CDID is 39x39 pixels. It should be noted that a smaller size of the search area leads to decreasing of noise suppression for homogeneous regions.

2) Increasing a sliding step to process every next reference block (patch). The default value in CDID is 3. It may be increased to 5 without drastic reducing of denoising efficiency. Maximal possible value should not exceed the size of patches. Decreasing the sliding step from 3 to 1 may lead to some increasing of denoising efficiency (in average on 0.1÷0.2 dB) with increasing a processing time up to 9 times.

3) Decreasing a block size leads in proportional decreasing of computational complexity. Default block size in CDID is 8x8. A smaller block size results in decreasing noise suppression efficiency for homogeneous regions, long edges and self-similar textures.

4) Decreasing a maximum number of similar blocks (maximum size of the 3rd dimension in the group) also leads to proportional decreasing computational complexity. Default value in CDID is 32. Decreasing this value to 8 decreases denoising efficiency for homogeneous regions and slightly increase denoising efficiency for textures and fine details.

## 5. Conclusion

A new class of non-iterative and iterative complex domain filters based on group-wise sparsity in complex domain is developed. This sparsity is based on the techniques implemented in Block-Matching 3D filtering (BM3D) and 3D/4D High-Order Singular Decomposition (HOSVD) developed for the spectrum analysis and filtering. The efficiency of the proposed algorithms is demonstrated by thoroughly simulation tests. In the paper we generalized these algorithms and implemented in MATLAB. One of the algorithms, the three step iterative ImRe-BM3D IT, is recognized as the most efficient for various test-images. In particular, it is shown, that in comparison versus the state-of-art algorithms SpInPhase and WFT this algorithm is the most precise for all complex-valued test images. The MATLAB toolbox for complex-domain (phase/amplitude) denoising is publicly available on http://www.cs.tut.fi/sgn/imaging/sparse/cdid.zip.

## 6. Acknowledgement

This work is supported by the Academy of Finland, project no. 138207, 2015-2019.